\documentclass[letterpaper, 10 pt, conference]{ieeeconf}  % Comment this line out if you need a4paper

\IEEEoverridecommandlockouts                              % This command is only needed if 
\usepackage{xcolor}
\usepackage{amsmath,amssymb,amsfonts}
\usepackage{booktabs}
\usepackage[linesnumbered,ruled,vlined]{algorithm2e}
\usepackage{subcaption} % For subfigures
\usepackage{nicefrac}       % compact symbols for 1/2, etc.
\usepackage{microtype}      % microtypography
\usepackage{cleveref}       % smart cross-referencing
\usepackage{graphicx}
\usepackage{balance}
\usepackage{fancyhdr}

\fancypagestyle{firstpage}
{
    \fancyhf{}

    \fancyhead[L]{\small \centering \noindent Paper accepted at the 2nd Workshop on Human-aligned Reinforcement Learning for Autonomous Agents and Robots HARL,\\at the IEEE International Conference on Robotics and Automation ICRA, Yokohama, Japan, 2024. 
    }

}

\title{\LARGE \bf
\BlankLine
Contextual Affordances for Safe Exploration in Robotic Scenarios
}

\author{William Z. Ye, Eduardo B. Sandoval, Pamela Carreno-Medrano, and Francisco Cruz% <-this % stops a space
\thanks{William Z. Ye and Francisco Cruz are with the School of Computer Science and Engineering, University of New South Wales, Sydney, NSW, Australia.}%
\thanks{Eduardo B. Sandoval is with the School of Art \& Design, Creative Robotics Lab, University of New South Wales, Sydney, NSW, Australia.}%
\thanks{Pamela Carreno-Medrano is with the Faculty of Engineering, Monash University, Australia.}%
}

\begin{document}

\maketitle
%\thispagestyle{empty}
%\pagestyle{empty}

%%%%%%%%%%%%%%%%%%%%%%%%%%%%%%%%%%%%%%%%%%%%%%%%%%%%%%%%%%%%%%%%%%%%%%%%%%%%%%%%
\begin{abstract}
Robotics has been a popular field of research in the past few decades, with much success in industrial applications such as manufacturing and logistics. This success is led by clearly defined use cases and controlled operating environments. However, robotics has yet to make a large impact in domestic settings. This is due in part to the difficulty and complexity of designing mass-manufactured robots that can succeed in the variety of homes and environments that humans live in and that can operate safely in close proximity to humans. This paper explores the use of contextual affordances to enable safe exploration and learning in robotic scenarios targeted in the home. In particular, we propose a simple state representation that allows us to extend contextual affordances to larger state spaces and showcase how affordances can improve the success and convergence rate of a reinforcement learning algorithm in simulation. Our results suggest that after further iterations, it is possible to consider the implementation of this approach in a real robot manipulator. Furthermore, in the long term, this work could be the foundation for future explorations of human-robot interactions in complex domestic environments. This could be possible once state-of-the-art robot manipulators achieve the required level of dexterity for the described affordances in this paper.
\end{abstract}

\thispagestyle{firstpage}

%%%%%%%%%%%%%%%%%%%%%%%%%%%%%%%%%%%%%%%%%%%%%%%%%%%%%%%%%%%%%%%%%%%%%%%%%%%%%%%%
\section{Introduction}
Current robots which operate in domestic settings have narrow use cases such as autonomous vacuum cleaners or coffee machines. These robots are designed to perform specific tasks and are not expected to interact with humans or learn to conduct new tasks actively~\cite{taniguchi2023world}. 
However, in a naturalistic domestic context, humans are around service robots and they can passively and actively interact with the robots~\cite{cruz2017agent}. 
Furthermore, humans can interfere with the robot's tasks and even get hurt if robots cannot manage to navigate the environment properly. For robots to be viable in a domestic setting, they must learn, recognise and meaningfully interact with humans whilst applying an understanding of social norms. 

Reinforcement Learning has proven to be a successful method for autonomous agents such as robots to learn desired behaviours based on rewarding and punishing actions. The algorithm often aims to balance agent exploration and exploitation of previous knowledge, whereby continual training leads to emergent strategy and sophisticated tool use and coordination~\cite{baker2020emergent}.

However, reinforcement learning is inherently an exploratory algorithm and whilst very effective in video games or in simulation where agents can make mistakes and experience many training iterations in parallel, RL becomes infeasible when considering real-world use cases where mistakes can be expensive or even dangerous. Hence we attempt to bolster the agent's knowledge base by using affordances as a baseline understanding of the world.

Affordances~\cite{gibson1979} are often aligned with the basic cognitive skills required to achieve an understanding of the context described~\cite{jamone2016affordances}. Intuitively humans learn these skills by acquiring knowledge of the affordances through an interaction with the environment surrounding them. This knowledge is then used to guide future actions and interactions with the world~\cite{cruz2016learning}. Similarly, we attempt to represent the contextual information of the environment such that a robot's actions can be better informed and thus reduce the time to learn new tasks as well as reduce the risk of interaction with other agents.

In this paper, we aim to develop an early exploration of contextual affordances for safe explorations in robotic scenarios that are familiar to humans too (domestic table with ordinary objects). Our results are promising and we expect to further develop this approach by implementing our findings in a robot embodiment shortly.

\section{Related Work}

% Affordances in general
The term affordance was initially introduced as an ecological proposition by~\cite{gibson1966} that characterises what actions are ``afforded'' to a certain agent (e.g. a human or a robot) in a given environment and what the consequences of these actions are for this agent. 

% Affordances in robotics
Gibson's definition of affordances has subsequently been adapted in many fields, including robotics. In robotics, affordances are often defined as relationships between an agent, object and effect~\cite{sahin2007} and play an important role in bridging robot perception to robot action, particularly during planning and control ~\cite{min2016affordance}. The optimal goal of modelling these relations among target objects, actions, and effects is for the robot to achieve human-like generalisation capabilities. According to~\cite{ardon2020affordances} the different approaches to affordances in robotics can be grouped into three main groups based on what they assume is known about these (object, action, effect) relations. The first group of approaches assume the robot has seen the required affordance relation before and can access it directly from a database of affordances built offline (e.g.~\cite{diana2013deformable},~\cite{price2016affordance}). The second group assumes the robot has partial a priori knowledge about the required affordance relationship since it has seen something similar in the past (e.g.~\cite{montesano2009learning},~\cite{griffith2012behavior}). The third and last group of approaches assume no a priori knowledge and thus the robot is required to learn the affordances relations via interactions with the environment or by applying heuristic rules on the perceived objects (e.g.~\cite{koppula2016anticipatory}).

Beyond their application to define object functionalities, affordances can also be seen as a way of characterising an agent's state in an action-oriented fashion, which can potentially lead to better generalisation and reduced learning time~\cite{yang2023recent}. This is of particular interest in reinforcement learning applications where an agent or robot learns to perform a task by interacting with its environment. Although RL has been demonstrated to be a powerful and useful learning approach (e.g. ~\cite{ibarz2021how}), its effectiveness is limited by the large number of interactions with the environment and the extensive computational resources required during training. This issue is further exacerbated in tasks with larger, high-dimensional state spaces. In this context, affordances can help limit the number of possible actions within a given state and thus help alleviate the computational and sampling complexity of RL learning algorithms~\cite{cruz2014improving}.

\section{Method}

\subsection{Preliminaries}
In a reinforcement learning problem, a decision-making agent (a robot in our case) interacts with its environment to learn an optimal sequence of actions that maximise its expected long-term return~\cite{sutton2018reinforcement}. This interactive learning process can be formalised using the Markov Decision Process (MPD) framework. An MDP is a tuple $(S, A, T, \gamma, r)$, where $S$ and $A$ denote the set of possible states and actions, $r: S \times A \rightarrow \mathbb{R}$ is the real-valued reward function, $T: S \times \mathcal{A} \times S \rightarrow \left[0,1\right]$ captures the environment's transition dynamics, mapping state-action pairs to a distribution over next states, and $\gamma$ is a discount factor that determines the relative importance between immediate and future rewards. For simplicity, we assume that the MDP's state and action spaces are finite.

At each time step, the robot observes a state $s_t \in S$ and takes an action $a_t \in A$ drawn from a policy $\pi: S \times A \rightarrow [0,1]$, which specifies the probability of selecting an action $a_t$ at a given state $s_t$ with $\sum_{a^t \in A} \pi(s^t,a^t) = 1$. Then, the robot receives a reward $r(s_t, a_t)$ and enters the next state $s_{t+1} \in S$ with probability $T(s_{t+1}|s_t, a_t)$. The value of taking an action $a$ in state $s$ under a policy $\pi$ is defined as: $Q_\pi(s,a) = \mathbb{E}\,[{\sum^\infty_{t=0}\gamma^t r(s_t, a_t)|s_t=s, a_t=a \; \forall t}]$. The goal of the agent is to find the optimal policy $\pi^*=\arg\max_\pi Q_\pi(s, a)$. If the model of the MDP, consisting of $r$ and $T$, is not given, the action-state value function $Q_\pi(s, a)$ can be estimated from experience.

\subsection{Proposed Approach}

We consider the problem where a robot must learn an optimal policy in settings where the set of actions available to the robot at a given state is limited ($A_s \subset A$) either because some actions are considered to be unsafe or not affordable. Specifically, consider the domestic cleaning scenario first introduced in~\cite{cruz2016} in which a robotic arm is tasked with cleaning a table. A single cup on the table impedes the ability of the robotic arm to clean the table freely. If the robotic arm were to clean the table without moving the cup first, it is likely the cup would break (unsafe/undesirable action). Similarly, if the robot already holds the cup, the cleaning action is not afforded at this current state since the robot's gripper is otherwise occupied.

\begin{figure}
    \centering
    \includegraphics[width=\columnwidth]{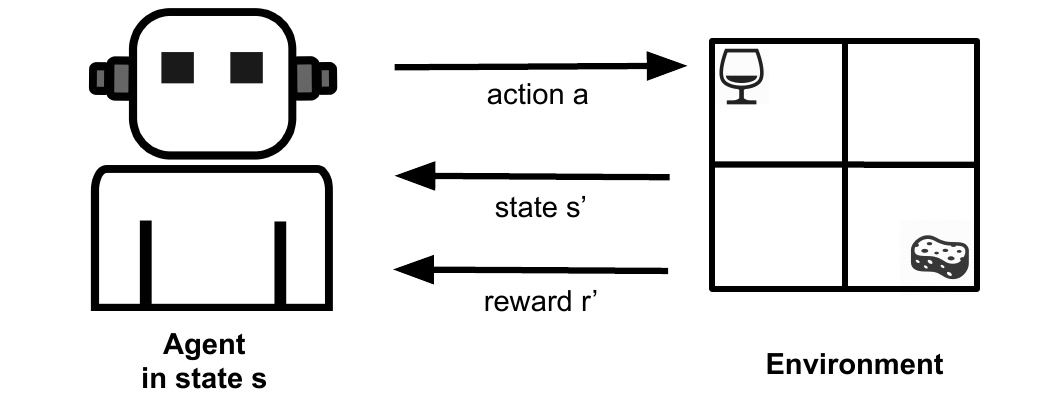}
    \caption{Simulated domestic cleaning scenario}
    \label{fig:RL}
\end{figure}

In this paper, we use contextual affordances~\cite{cruz2016learning} to provide the robotic arm with knowledge about actions that will lead to undesirable states (e.g. broken cup) or failed states (e.g. gripper already occupied) from which it is not possible to reach the final objective (e.g. a clean table). However, instead of considering small state spaces (i.e. experiments in~\cite{cruz2016} were limited to a table with 3 possible locations), we propose a state representation that allows the robotic arm to learn a suitable cleaning policy in larger, 2-dimensional table representations, as shown in Figure~\ref{fig:RL}.

\subsection{State and Action Representations}
The table is represented as a grid of cells, where each cell has a state that can be either dirty or clean. No positional information is assigned to each cell, however, on creation of the environment the order of initialisation of each cell determines its position. We also represent two objects, a sponge and a cup, with each object containing a positional encoding expressed as a Cartesian coordinate to represent its location on the table. Finally, the agent is represented as a robotic arm with a positional encoding as well as a held object encoding which can be one of $<$None, Cup, Sponge$>$.

\begin{table}[ht]
	\caption{State space representation}
	\centering
	    \begin{tabular}{@{}p{1.7cm}p{2.5cm}l@{}}
		\toprule
		Property 		& Description     		& Encoding \\
		\midrule
		Width		 		&Table width  		& Int \\
		Height	 		& Table height 		& Int \\
		Cells		 		& List of cells   		& Cell($<$Clean, Dirty$>$) \\
		Cup 				& Cup position 		& Cup(Int, Int) \\
		Sponge  		& Sponge position 	& Sponge(Int, Int) \\
		Robot 	& Agent           		& Robot(Int, Int, Object) \\
		\bottomrule
	\end{tabular}
	\label{tab:statemachine}
\end{table}

Table~\ref{tab:statemachine} depicts the proposed state representation. We also augment the action space to include seven possible actions in total (see Table~\ref{tab:actions}). We note that compared to~\cite{cruz2016}, we introduced a third cell to replace the ``home" action. Additionally, for problems with only a single row of cells, we disable the Go Up and Go Down actions to minimize the action space of the agent.

\begin{table}[ht]
	\caption{Action space}
	\centering
	\begin{tabular}{@{}lp{6cm}@{}}
		\toprule
		Action      & Description \\
		\midrule
		Pick Up	    & Pick up object at position of the robotic arm \\
		Put Down    & Put down object at position of the robotic arm \\
		Clean       & Clean the cell at the position of the robotic arm \\
		Go Left     & Move the robotic arm one cell left \\
		Go Right    & Move the robotic arm one cell right \\
		Go Up       & Move the robotic arm one cell up \\
		Go Down     & Move the robotic arm one cell down \\
		\bottomrule
	\end{tabular}
	\label{tab:actions}
\end{table}

\section{Experimental Setup}

The exploration cost of RL algorithms exponentially increases with the size of the state space. However, a good representation for the problem at hand can increase the success of RL algorithms in practice~\cite{du2020good}, thus we encode the proposed state representation as a single integer through the following process. 
The bottom n bits represent n-cells and whether they are Clean (0) or Dirty (1). The next $\text{ceil}(\log_{2}\text{n\_cells})$ bits then encode the position of the arm, and similarly, we encode the position of the sponge and the cup in the same way. Finally, the highest 2 bits are used to encode which object is held by the Robotic Arm (0 - None, 1 - Cup, 2 - Sponge).

The width and height of the table are required to encode and decode the state representation and thus the encoding is not transferable between different scenarios. This encoding scheme was chosen to allow us to scale the representation to larger table dimensions, and as a result, the encoding has extra bits that carry no information. For example, to represent a positional encoding we use 2 bits to represent 3 cells, despite the 2 bits being able to carry information for 4 positions. With this encoding, a 3-cell table environment similar to the one proposed in~\cite{cruz2016} has 1536 possible states versus the more tightly controlled 45 states. The number of states increases exponentially for each cell added to the table representation and if we consider the (state, action) pairs, the state representation grows even larger for table representations that are 2-dimensional (see Table~\ref{tab:statespace}). 

\begin{table}[ht]
	\caption{Table State Space Encoding Size}
	\centering
	\begin{tabular}{@{}lll@{}}
		\toprule
		Dimensions & Total states & Total (state, action) pairs\\
		\midrule
		3 x 1	   & 1,536    & 7680\\
		4 x 1    & 3,072    & 15,360\\
		5 x 1    & 49,152   & 245,760\\
		6 x 1    & 98,304   & 491,520\\
		7 x 1    & 196,608  & 983,040\\
    2 x 2    & 3,072    & 21,504\\
    3 x 2    & 98,304   & 688,128\\
		\bottomrule
	\end{tabular}
	\label{tab:statespace}
\end{table}

A SARSA RL algorithm is used to train the robotic arm to learn the domestic cleaning task. The SARSA algorithm is an on-policy method where the agent learns the state-action value function according to actions derived from the current policy~\cite{rummery1994}. We update the state-action value function according to: $Q(s_, a_t) \leftarrow Q(s_t, a_t) + \alpha [r_{t+1} + \gamma Q(s_{t+1}, a_{t+1}) - Q(s_t, a_t)]$ where $s_t$ and and $a_t$ represent the current state and action, whilst $s_{t+1}$ and $a_{t+1}$ represent the next state and action respectively, and $\alpha$ is the learning rate. 
The reward function delivers $+1$ as the reward if we reach the single final state and $-1$ if we reach a failed state. In both scenarios, this will end the current learning iteration. In any other situation the reward
function applied a small negative reward of $-0.01$ to encourage shorter paths, depicted in Equation~\ref{eq:reward}.

\begin{equation}
  \label{eq:reward}
  r(s, a) = 
  \begin{cases}
    1.0   & state\ is\ final  \\
    -1.0  & state\ is\ failed \\
    -0.01 & otherwise
    \end{cases}
\end{equation}

For our experiment, we use $\alpha = 0.3$ and $\gamma = 0.9$. During learning, the robotic arm chooses an action according to the $\epsilon$-greedy method demonstrated by Algorithm~\ref{alg:action_selection} with an $\epsilon = 0.1$.

\begin{algorithm}
  \KwData{state $s_t$}
  \KwResult{action $a_t$}
  \If{$rand(0, 1) < \epsilon$}{
    $a_t = random(A)$
  }
  \Else{
    $a_t = max(Q(s_t, \cdot))$
  }
  \caption{$\epsilon$-greedy action selection (standard RL)}
  \label{alg:action_selection}
\end{algorithm}

% \subsection{Deep Contextual Affordance Learning}
% Using a previous run of the SARSA RL algorithm, we save a dataset consisting of the $(state, action) \rightarrow
% (failed, next\_state)$ to be used as the training data. Since the state representation contains the position of the
% object as well, then this data represents the contextual affordance $f(state, object, action) \rightarrow effect$. 

% We use a simple feedforward architecture~\cite{sazli2006} with 2 fully connected hidden-layers with 256 and 64 nodes
% respectively depicted in Figure~\ref{fig:fig1}. We chose a larger network in comparison to~\citet{cruz2016} due to the
% size of the input data and the higher number of states represented.

% \begin{figure}[h]
% 	\centering
%   \includegraphics[width=0.7\textwidth]{images/mlp.jpg}
% 	\caption{Feedforward Network Architecture}
% 	\label{fig:fig1}
% \end{figure}

% \subsection{Attention-based Affordance Learning}
To adapt an attention-based model we implement the architecture described by~\cite{vaswani2023attention} however
implementing only the decoder half of the network. Our architecture consists of 6 Transformer Layers and we adapt a model used for learning character-based text encodings to our problem.

%We modify the dataset such that the state, action pairs are serialized providing a history of each training run and set the vocabulary to the state and action representations. We additionally add one more state to represent a failed run and present the tokens to the attention-based model through a masked-attention head, preventing the model from viewing
% information after the token it is looking at. Figure~\ref{fig:transformer} depicts the implemented architecture and
% notably differs from~\citet{vaswani2023attention} by the lack of cross-attention with an encoder block.

% \begin{figure}[h]
% 	\centering
%   \includegraphics[width=0.2\textwidth]{images/transformer.png}
% 	\caption{Attention-based Architecture}
% 	\label{fig:transformer}
% \end{figure}

%\subsection{Reinforcement Learning with Contextual Affordances}
We integrate the contextual affordances during learning by generating $(state, action) \rightarrow failed\_state$ observations with an accuracy of $90\%$. 
Using this information, during exploration, we define the subset of actions $A_s$ as the set of actions that are afforded or considered safe and modify the $\epsilon-$greedy action selection method as seen in Algorithm~\ref{alg:action_selection_afford} by only allowing exploration if the next action is deemed not dangerous. 

\begin{algorithm}[ht]
  \KwData{state $s_t$}
  \KwResult{action $a_t$}
  $A_s = \text{affordances}(s_t, \forall a \in A)$

  \If{$rand(0, 1) < \epsilon$}{
    $a_t = random(A_s)$
  }
  \Else{
    $a_t = max(Q(s_t, \cdot))$
  }
  \caption{$\epsilon$-greedy action selection with affordances}
  \label{alg:action_selection_afford}
\end{algorithm}

\section{Results}

% Figure~\ref{fig:3_1_avg_reward} shows the comparison of the average collected rewards for the classical RL algorithm
% compared to the RL algorithm using the feedforward model to generate $A_s$. In this manner we find that the agent learns
% incredibly quickly and rapidly begins to complete only successful runs, increasing the number of successful runs from
% 50,086 to 78,952 in total out of the 100,000 training iterations.

% \begin{figure}[h]
% 	\centering
%   \includegraphics[width=0.5\textwidth]{images/avg_reward_3_1.png}
% 	\caption{Average Collected Rewards with and without Contextual Affordances: Table State 3 x 1}
% 	\label{fig:3_1_avg_reward}
% \end{figure}

\begin{figure*}[ht]
  \centering
  \begin{subfigure}{0.4\linewidth}
      \centering
      \includegraphics[width=\linewidth]{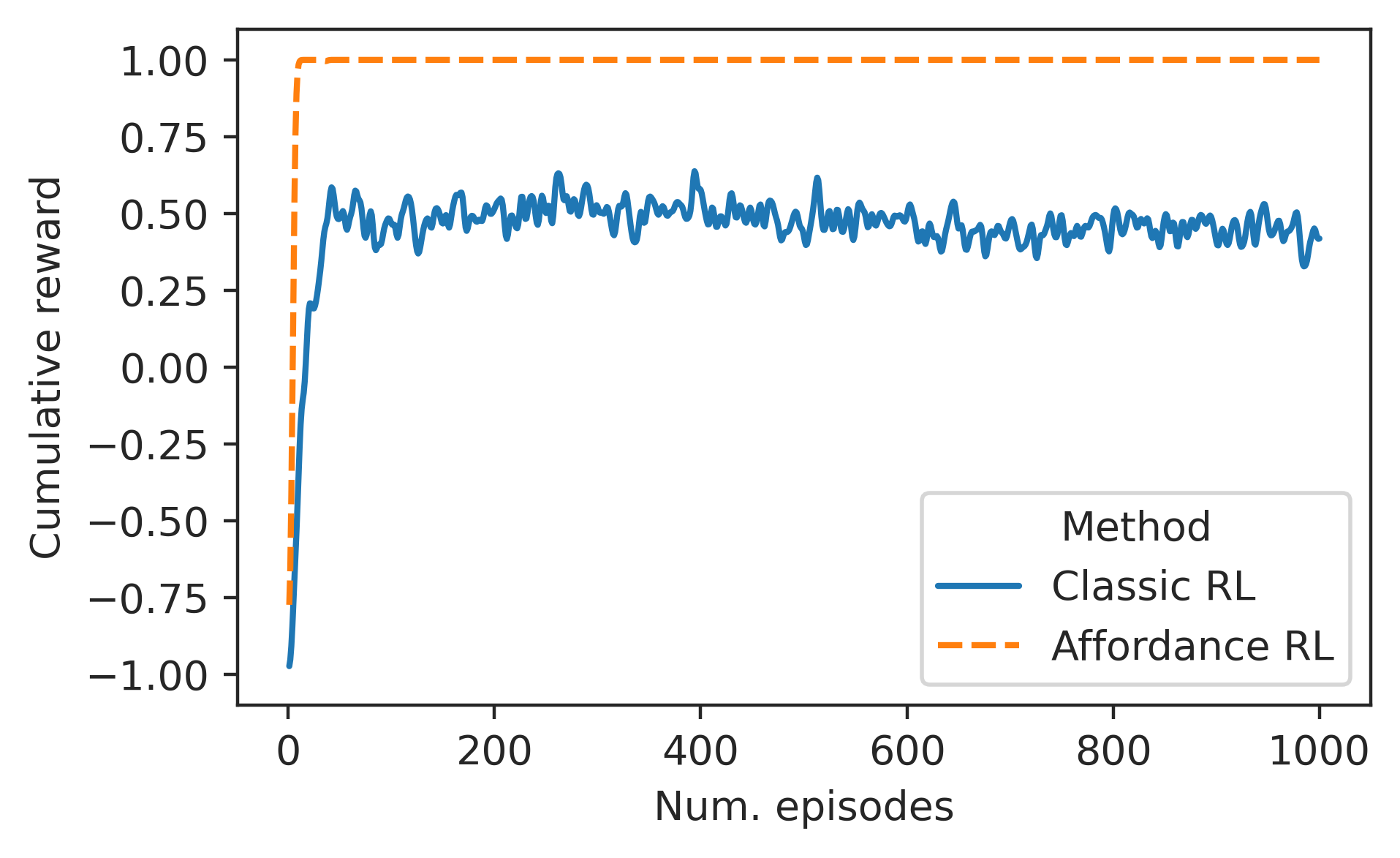}
      \caption{$3 \times 1$ Table}
  \end{subfigure}
  \begin{subfigure}{0.4\linewidth}
      \centering
      \includegraphics[width=\linewidth]{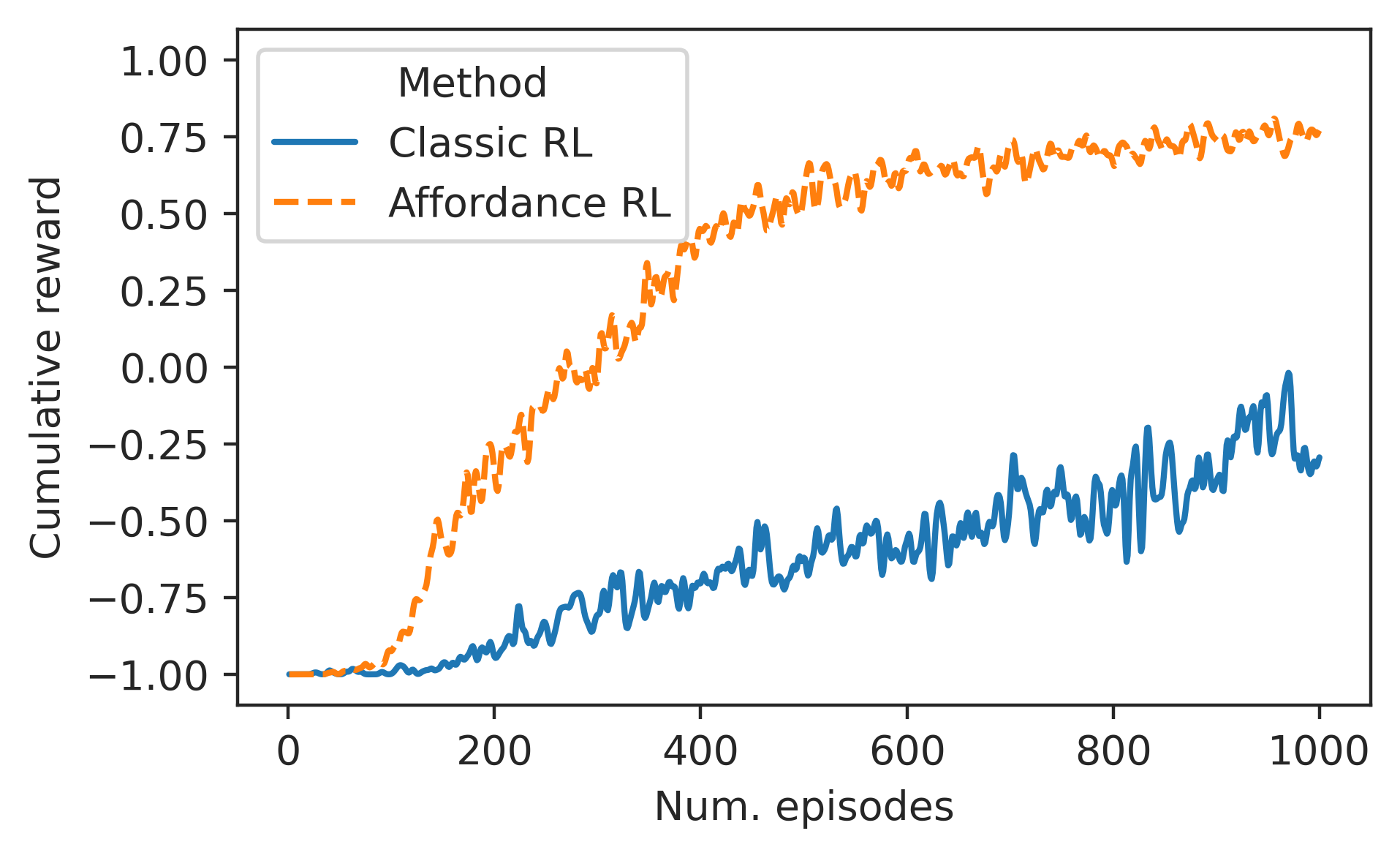}
      \caption{$6 \times 1$ Table}
  \end{subfigure}
  \begin{subfigure}{0.4\linewidth}
      \centering
      \includegraphics[width=\linewidth]{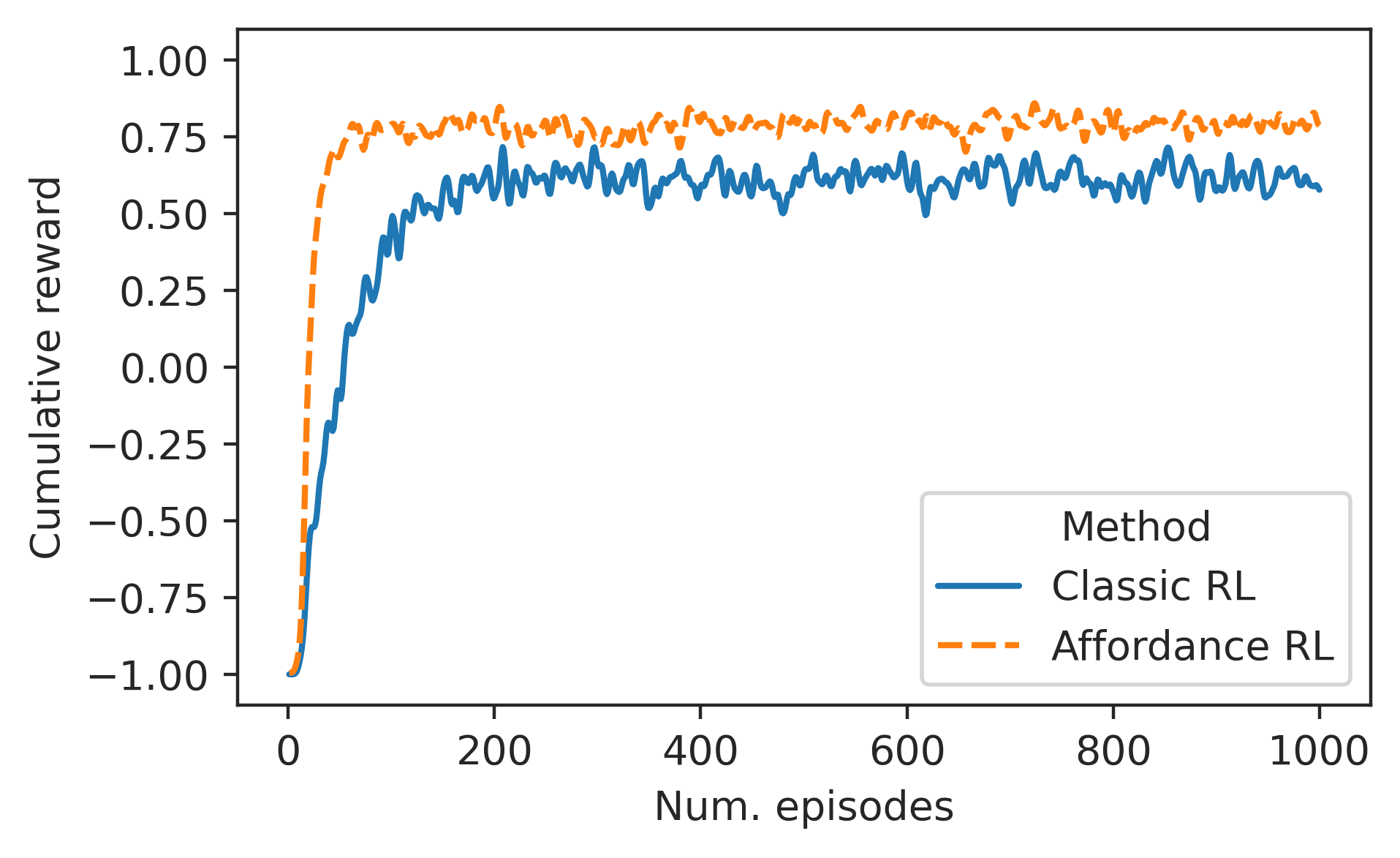}
      \caption{$2 \times 2$ Table}
  \end{subfigure}
  \begin{subfigure}{0.4\linewidth}
      \centering
      \includegraphics[width=\linewidth]{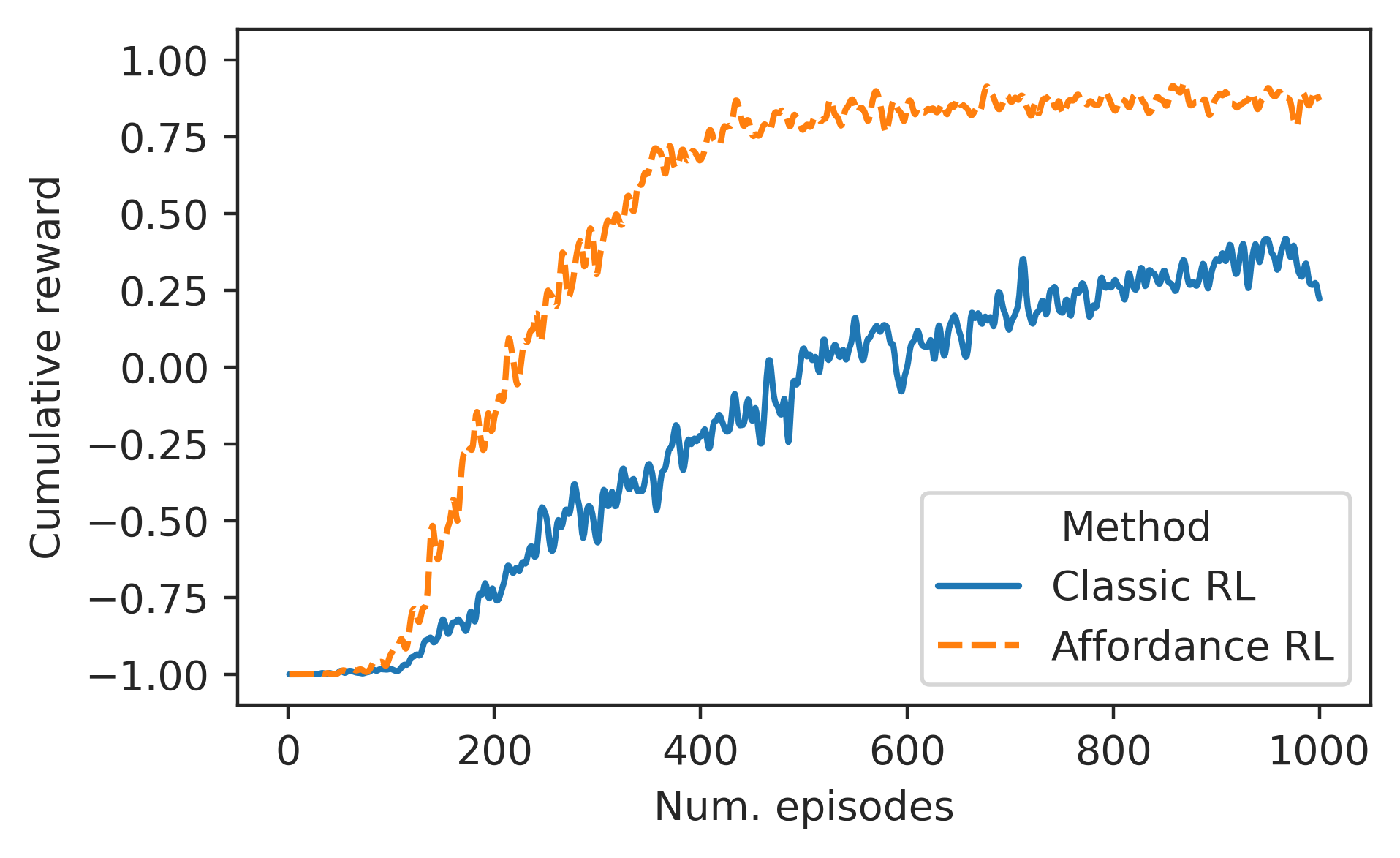}
      \caption{$3 \times 2$ Table}
  \end{subfigure}
  \caption{Average collected rewards during training with (orange dashed line) and without (blue solid line) contextual affordances.}
  \label{fig:avg_reward_4_1_5_1}
\end{figure*}

Figure~\ref{fig:avg_reward_4_1_5_1} shows the comparison of the average cumulative rewards across training runs for different table dimensions. For each table dimension, a SARSA algorithm was trained for a total of $ 100,000$ iterations. As mentioned previously, these results used a subset $A_s$ that was generated synthetically, however, we can see that by avoiding failed states altogether, the agent is much better at learning the task at hand across all dimensions. The $6 \times 1$ table environment demonstrates an interesting point for the reinforcement learning algorithm. Within the 100,000 training iterations (1000 episodes x 100 steps per episode), the classical RL algorithm was unable to learn the domestic cleaning task. With the affordances incorporated, however, the agent was able to reach an average collected reward of $0.75$. 

We also note that although the contextual affordances narrow down the search space of the robot during training when compared to the classical approach, more training iterations are required in order to converge to a policy that achieves the maximum expected return as the dimensionality of the state space increases. For instance, for the $3 \times 1$ table environment, we find that the agent learns quickly (less than 100 episodes) and rapidly begins to complete only successful runs. However, for the other environments, even after 100,000 training iterations the robotic arm seems to incur in additional costs and can only achieve an average collected reward of $0.75$.

% \begin{table}[h]
% 	\caption{Average rewards in evaluation}
% 	\centering
% 	\begin{tabular}{@{}lll@{}}
% 		\toprule
% 		Dimensions & Classic RL & Affordances RL\\
% 		\midrule
% 		3 x 1	   & 0.57 $\pm$ 0.69  & 0.86 $\pm$ 0.06\\
% 		4 x 1    &  -0.01 $\pm$ 0.94  & 0.20 $\pm$ 0.88 \\
% 		5 x 1    & -0.65 $\pm$ 0.72  & 0.004 $\pm$ 0.91\\
% 		6 x 1    & -0.83 $\pm$ 0.53   & -0.11 $\pm$ 0.89\\
% 		7 x 1    & -0.89 $\pm$ 0.42  & -0.93 $\pm$ 0.33\\
%     2 x 2    & 0.28 $\pm$ 0.87 & 0.71 $\pm$ 0.48\\
%     3 x 2    & -0.13 $\pm$ 0.92 & 0.49 $\pm$ 0.69\\
% 		\bottomrule
% 	\end{tabular}
% 	\label{tab:statespace}
% \end{table}

\section{Discussion}
\label{Discussion}

As the results could make sense in the context of reinforcement learning, further iterations are required to transfer them into an implementation involving human-robot interactions in the real world. In order to make possible a future implementation, we suggest the following conditions should be met:

\begin{enumerate}
    \item Implement the results using an appropriated robot embodiment. It is required to choose a robot arm capable of accomplishing the proposed task used in the training and simulation described in this paper. In other words, we require a robot with a high level of dexterity which potentially is still not in the market.
    \item The design of an appropriated feasible task as the one simulated and described in this paper. The task proposed for the training of this algorithm presents different layers of complexity in the real world. Certainly, it is required to test the feasibility and technology readiness of state-of-the-art robots to implement an effective human-robot interaction in domestic environments. 
    \item The long-term implications of this project require involving a human in the interaction related to the proposed task and considering the safety and security factors. Maintaining human safe when interacting with robots manipulating objects in the physical world is the highest priority in any human-robot interaction.
\end{enumerate}

\section{Conclusions and Future Work}

This work demonstrates how introducing information during the action selection process for reinforcement learning can dramatically improve the convergence of the model. Ultimately a robot in the real world would operate off of sensors such as RGB-D cameras and microphones (among other sensors) and thus applying affordance learning in a physics simulator for the same domestic cleaning task would work towards this goal. Furthermore, the trained model could then be transferable to a real-world robotic arm. However, the design of the robotic scenario has to be refined in order to transfer our model to an acting robot interacting with complex affordances in the real world.

For both of the tasks described above, it would be beneficial to move from learning on discrete state spaces such as the state machine described in this report. As a robot moves around the environment, it will be continuously receiving streams of data from all its sensors and thus an ML model capable of learning continuously on this stream of data would likely benefit the learning capabilities of the agent. In the long term, it would likely be the mainstream approach used in robots providing a service in complex domestic environments interacting with everyday objects.

%\subsection{Future work}

As mentioned in the section~\ref{Discussion}, the following challenge, after testing our proposed system, is to implement it in a more feasible and communicable way to be understood by other researchers coming from different fields. Firstly, a visual simulation using the previous training is required. The affordances defined in this paper should match as closely as possible with the affordances in the real world. Secondly, transfer the optimal policies proposed in this paper in a real robot characterising the movements in a given environment considering the consequences of these actions for the agent and potentially with other agents interacting (humans) in the same environment. Particularly, we should focus on the unsafe or not affordable actions impacting significantly the human-robot interaction.
%% This section was initially prepared using BibTeX.  The .bbl file was
%% placed here later

%\balance
%\bibliography{references}
%\bibliographystyle{named}

\bibliographystyle{IEEEtran} 
\balance
\bibliography{mybibfile}


\end{document}